\documentclass[conference]{IEEEtran}
\IEEEoverridecommandlockouts
\usepackage{cite}
\usepackage{amsmath,amssymb,amsfonts}
\usepackage{algorithmic}
\usepackage{graphicx}
\usepackage{textcomp}
\usepackage{xcolor}
\usepackage{amsmath,lipsum}
\usepackage[ruled]{algorithm2e}
\usepackage{graphicx}
\usepackage{float}
\usepackage{subfigure}
\usepackage{multirow}
\usepackage{multicol}
\usepackage{arydshln}
\usepackage{cite}
\usepackage{stfloats}
\usepackage{booktabs}
\usepackage{caption}

\def\BibTeX{{\rm B\kern-.05em{\sc i\kern-.025em b}\kern-.08em
    T\kern-.1667em\lower.7ex\hbox{E}\kern-.125emX}}
\begin{document}

\title{Attention-based Feature Compression for CNN Inference Offloading in Edge Computing\\ }
\author{\IEEEauthorblockN{Nan~Li, Alexandros~Iosifidis and Qi~Zhang }
\IEEEauthorblockA{DIGIT, Department of Electrical and Computer Engineering, Aarhus University.\\
Email: \{linan, ai, qz\}@ece.au.dk}}
\maketitle

\begin{abstract}
This paper studies the computational offloading of CNN inference in device-edge co-inference systems. Inspired by the emerging paradigm semantic communication, we propose a novel autoencoder-based CNN architecture (AECNN), for effective feature extraction at end-device. We design a feature compression module based on the
channel attention method in CNN, to compress the intermediate data by selecting the most important features. To further reduce communication overhead, we can use entropy encoding to remove the statistical redundancy in the compressed data. At the receiver, we design a lightweight decoder to reconstruct the intermediate data through learning from the received compressed data to improve accuracy. To fasten the convergence, we use a step-by-step approach to train the neural networks obtained based on ResNet-50 architecture. Experimental results show that AECNN can compress the intermediate data by more than $256 \times$ with only about 4\% accuracy loss, which outperforms the state-of-the-art work, BottleNet++.
Compared to offloading inference task directly to edge server, AECNN can complete inference task earlier, in particular, under poor wireless channel condition, which highlights the
effectiveness of AECNN in guaranteeing higher
accuracy within time constraint.
\end{abstract}


\section{Introduction}
Convolutional neural networks (CNNs) have recently achieved remarkable success in many IoT applications. To enable reliable services and intelligent control, not only high reliability but also
low latency is crucial for time-critical IoT applications \cite{NanGRL2022}. However, the large amount of weight parameters and high computation overhead of CNN hinder the \textit{local computing} on resource-limited IoT devices.

To address this issue, a common approach is to compress and prune CNN topology and thereby reducing the computational operations on IoT device. However, over-compressing CNNs may cause severe accuracy degradation. Another approach is \textit{full offloading}, which allows IoT devices to offload computational tasks directly to edge servers (ESs) in their proximity \cite{008}. However,
the fluctuations in communication time caused by stochastic wireless channel states may result in the inference missing the service deadline \cite{Nan2022}. These limitations have driven the development of other alternatives, among which \textit{device-edge co-inference} is a promising approach that can strike a balance between communication and computation \cite{BottleNet++}. In this approach, a CNN model is split between an IoT device and an ES, and the IoT device sends the intermediate data to ES for further processing. Note that the intermediate data is a tensor consisting of numerous channels and each channel of the tensor is a two-dimensional feature.

Most of the existing works on device-edge co-inference mainly focus on model splitting, while less attention has been paid to the compression of intermediate tensor \cite{BottleNet++}. In general, a well-trained CNN usually has more redundant features than needed to perform an inference task \cite{Han_2020_CVPR}, and not all of these features play the same role therein, i.e., different features have different importance for making predictions \cite{Woo_2018_ECCV}. Therefore, under poor channel condition, it is feasible to prune some of the less important features to reduce the communication overhead, thereby meeting the deadline. This is aligned with the emerging paradigm of \textit{semantic communication}, which aims at extracting the ``meaning" of the information to be sent at a transmitter, and interpreting the received semantic information successfully at a receiver~\cite{Luo2022}. A semantic communication system typically consists of a semantic encoder and a semantic decoder, where the encoder plays the role of encoding large raw data (e.g., high-dimensional intermediate tensor) into a more compact data representation (e.g., reduced-dimensional tensor) and the decoder is to complete the task using the received information.

Based on the motivation above, this paper proposes a novel autoencoder-based CNN architecture (AECNN) for computational offloading of CNN inference task in a device-edge co-inference system. Our contributions include: 1) We design a \textit{channel attention} (CA) module to quantify the importance of channels in the intermediate tensor, and use the statistics of channel importance to calculate the importance of each channel, which enables intermediate tensor compression by pruning the channels of lower importance. Moreover, we can use entropy encoding to remove the statistical redundancy in the compressed intermediate tensor, to further reduce the communication overhead; 2) We design a light-weight \textit{feature recovery} (FR) module that uses a CNN to learn and recover the intermediate tensor from the received compressed tensor, thereby improving inference accuracy; 3) We use a step-by-step approach to fasten the training of the resulting neural networks of ResNet-50 architecture \cite{He_2016_CVPR}. Experimental results show that AECNN achieves more than $256 \times$ compression for intermediate features with only about 4\% accuracy loss, which outperforms the state-of-the-art work, BottleNet++ \cite{BottleNet++}. Compared to \textit{full offloading}, AECNN can perform inference faster 
under poor wireless channel, which demonstrates that
AECNN can achieve higher accuracy within time constraints.

\section{Related work}\label{section:system_model}
Due to the large number of multiply-accumulate operations in deeper CNNs, resource-limited IoT devices are not feasible to perform inference within a stringent deadline. In addition, offloading an entire computation task to ES may be limited by uplink wireless bandwidth. To strike a balance between the computation and communication, \textit{device-edge co-inference} can be a good alternative to speed up CNN inference.
\begin{figure*}[]
    \centering
    \includegraphics[width=0.7\textwidth]{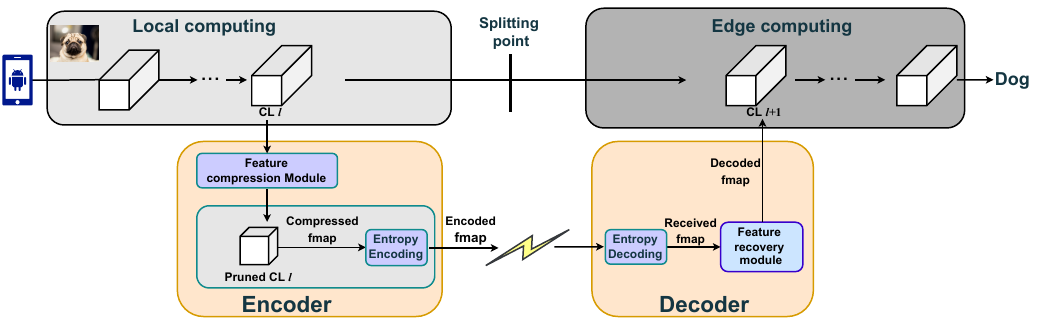}
    \caption{The proposed AECNN architecture in device-edge co-inference system.}
    \label{fig:AECNN}
        \vspace{-1mm}
\end{figure*}

\subsection{Dynamic model splitting}
Dynamic model splitting is an important topic in \textit{device-edge co-inference} systems. Kang et al. \cite{Neurosurgeon} proposed Neurosurgeon, which splits a CNN model based on the energy consumption and latency of each layer to reduce the end-to-end latency while satisfying energy and memory constraints. Li et al. \cite{EdgeAI} proposed Edgent, which splits a CNN model according to the available bandwidth to maximize the inference accuracy while satisfying application latency requirements. However, the latency gain achieved by these methods is limited. As the size of the intermediate tensor of the early layers may exceed the raw data (i.e., raw image), a CNN model can only be split in the latter layers to avoid too high communication overhead. However, this may cause excessive computation time on the IoT device.

\subsection{Feature compression}
To solve this issue, feature compression is used to compress the intermediate tensor, so that a CNN model can be split at the early layers. Ko et al. \cite{Ko2018} proposed to apply JPEG coding to compress the intermediate tensor. However, the redundant features in a well-trained CNN limit the latency gain achieved by the lossy compression method. Motivated by the fault-tolerant property of CNNs, Shao et al. \cite{BottleNet++} proposed BottleNet++, which compresses the intermediate tensor by using a CNN module to resize the feature dimension. However, directly resizing feature dimension may compromise the effective representation of the semantic information in the features and result in accuracy degradation.

\section{system model}\label{section:system_model}
In this paper, we study the computational offloading of CNN inference tasks
in a device-edge co-inference system, where a CNN model is split between an IoT device and an edge server, as shown in Fig. 1. We consider a CNN model $\mathcal{G}$ with $L$ convolutional layers (CLs) and denote them as $ \mathcal{L} = \left\{1, \cdots, L \right\}$. Let us assume that the CNN model is split at CL $l \in \mathcal{L}$, the IoT device can offload the intermediate tensor to the ES via a wireless channel after performing the local computation (i.e., from CL 1 to $l$), and then the ES performs the rest of the computation (i.e., from CL $l+1$ to $L$) and sends the inference results back to the IoT device. In the case of poor wireless channel states, it is not feasible for the IoT device to offload intermediate tensor directly to the ES because the large communication overhead will result in failure to meet deadline. As such, an autoencoder can be used to compress the intermediate tensor at a pre-defined compression ratio, to reduce communication time without compromising too much accuracy. We use a set $ \mathcal{V}=\left\{\mathcal{V}_1, \cdots, \mathcal{V}_k, \cdots, \mathcal{V}_K\right\}, \mathcal{V}_k > 1$ to denote all the pre-defined compression ratios.

\subsection{Communication time}
Computation offloading involves transmitting intermediate tensor and its inference result between IoT device and ES. Since the output of a CNN is very small, usually a number or few values representing the classification or detection result, the transmission delay of the feedback is negligible.

Let us assume that the output tensor of CL $l$ is $X \in \mathbb{R}^{C_l \times H_l \times W_l}$ of type float32. Its size in bytes is 
\begin{equation}
    S_l = 4\cdot C_l \cdot H_l \cdot W_l
\label{eq1}
\end{equation}
where $C_l$, $H_l$ and $W_l$ are the dimensions of the output tensor's channel, height and width, respectively.

We assume that the uplink transmission data rate from the IoT device to the ES is $r$. We also assume that the CNN model is split at CL $l$ and the compression ratio of the encoder is $\mathcal{V}_k$, then, the communication time of transmitting the compressed intermediate tensor is
\begin{equation}
t_{l,\mathcal{V}_k}^{\textit{com}} = 
    S_l/\left(\mathcal{V}_k \cdot  r\right).
\label{eq2}
\end{equation}

\subsection{Computation time}
In the device-edge co-inference system, the computation time consists of the CNN computation and feature encoding time on the IoT device, and the feature decoding and CNN computation time on the ES. Therefore, with the CNN model split at CL $l$ and the compression ratio of encoder $\mathcal{V}_k$, the total computation time can be expressed as
\begin{equation}
t_{l,\mathcal{V}_k}^{\textit{cmp}} = t_{l,\mathcal{V}_k}^{\textit{iot}} + t_{l,\mathcal{V}_k}^{\textit{enc}} + t_{l,\mathcal{V}_k}^{\textit{dec}}+ t_{l+1}^{\textit{es}},
\end{equation}
where $t_{l,\mathcal{V}_k}^{\textit{iot}}$ and $t_{l,\mathcal{V}_k}^{\textit{enc}}$ represent the time taken by IoT device to perform the computation from the input layer to CL $l$ and the entropy encoding of the feature map; $t_{l,\mathcal{V}_k }^{\textit{dec}}$ and $t_{l+1 }^{\textit{es}}$ represent the time taken by ES to perform the entropy decoding and computation from CL $l+1$ to the end. Note that the feature compression module is only used for training the pruned CL $l$ but not for inference, and the time to perform the computation of the light-weight FR module on ES is very small and negligible.

\subsection{Completion time of an inference task}
The total completion time of an inference task includes the CNN computation and feature encoding time on the IoT device, the communication time for transmitting the compressed intermediate tensor from IoT device to ES, and the
feature decoding and CNN computation time on the ES. Therefore, with the split at CL $l$ and compression ratio $\mathcal{V}_k$, the total completion time is
\begin{equation}
    t_{l,\mathcal{V}_k} = t_{l,\mathcal{V}_k}^{\textit{com}} + t_{l,\mathcal{V}_k}^{\textit{cmp}}.
\end{equation}

\section{Architecture of the proposed AECNN}
In this section, we first present an overview of our proposed AECNN architecture. Next we describe the structural components of our designed feature compression module in the encoder and how to compress the intermediate tensor. Then, we introduce the designed \textit{feature
recovery} (FR) module in the decoder. Finally, we present the training strategy of our proposed AECNN.

\subsection{Overview of the proposed AECNN architecture}
Fig. \ref{fig:AECNN} depicts the overall framework of our proposed AECNN, which consists of an encoder and a decoder. In the implementation of the encoder, a \textit{channel attention} (CA) module is designed to rank the channels by their importance for inference accuracy using statistical information of the channels’ importance. In this way, channels with low importance are pruned based on the pre-defined compression ratio $\mathcal{V}_k$. Finally, an entropy coding module can be applied to the remaining intermediate features to remove the statistic redundancy in the data. The decoder first uses the entropy decoding module to decode the received data, and then uses the designed FR module to recover the pruned features from the received features.
\subsection{Feature compression module}
The attention mechanism can effectively improve the classification performance of CNNs by enhancing the representation of features with more important information and suppressing unnecessary information interference \cite{Woo_2018_ECCV}. Channel attention used in CNN usually focuses on evaluating the importance of tensor's channels by paying attention to different channels of the tensor. For example, for CL $l$, each element of the channel attention map $Q_l \in \mathbb{R}^{C_l \times 1 \times 1}$ corresponds to a channel's weight of the output tensor $X_l \in \mathbb{R}^{C_l \times H_l \times W_l}$. As such, the channels with lower importance can be identified and removed, thereby reducing the size of the intermediate tensor and reducing the communication and computation time on the IoT device. Note that as IoT device knows which channels will be pruned for any pre-defined compression ratio, it only needs to compute the retained channels of the output tensor.
\begin{figure}
    \centering
    \includegraphics[width=0.38\textwidth]{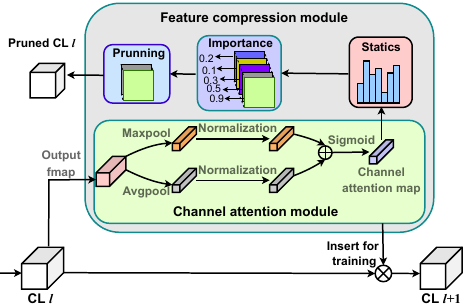}
    \caption{Feature compression module of the Encoder.}
    \label{fig:encoder}
        \vspace{-3mm}
\end{figure}

In the previous designs of channel attention \cite{Woo_2018_ECCV}, two fully-connected (FC) layers are used to handle the attention weight of the channels. However, this may introduce two drawbacks. First, the reduction of channel dimensionality for saving computation overhead may have side effects on the prediction of channel attention. Second, the learned channel attention by FC is intrinsically implicit, resulting in unknowable behavior of neuronal output. To address these issues, normalization can yield competition or cooperation relationships among channels, using fewer computation resource while providing more robust training performance \cite{Yang_2020_CVPR}. Motivated by the above, we design a CA module, i.e., a global max-pooling layer and a global average-pooling layer with normalization, and insert it into the original CNN model after the splitting point $l$, and then train the resulting network to generate the importance value of each channel, as shown in Fig. \ref{fig:encoder}. The detailed computational process of the proposed CA module is described in the following.

Since the calculation of the global avg-pooling layer and global max-pooling layer are similar, we take the avg-pooling layer as an example. The aggregated features after the avg-pooling layer can be represented as
\begin{equation}
    A_l = \textit{AvgPool} \left(X_l\right).
\end{equation}
where $A_l \in \mathbb{R}^{C_l \times 1 \times 1}$. 

The aggregated features $A_l$ is normalized as 
\begin{equation}
    \hat{A}_l = \frac{A_l-\mu}{\sqrt{\delta^2+ \epsilon}}
\end{equation}
where $\epsilon>0$ is a small constant, and $\mu$ and $\delta$ are the mean and the standard deviation of $A_l$, respectively.

Then, the normalized features are subjected to element-wise summation and \textit{sigmod} activation operation to generate the final channel attention map $Q_l$ as 
\begin{equation}
    Q_l = \textit{sigmod}\left( \hat{A}_l +  \hat{M}_l\right),
\end{equation}
where $\hat{M}_l$ is the normalized features of max-pooling layer.

Note that the generated channel attention weights vary depending on the input data (i.e., images), as shown in Fig. \ref{fig:attention} in the experimental results. To measure the importance of the channels, we use the statistical information found by element-wise averaging the weight in the channel attention map for all training data. The importance of channel $c$ for the intermediate tensor of CL $l$, $q_l^{c} \in Q_l$, can be calculated as
\begin{equation}
    q_l^{c} = \frac{1}{|D|}\sum_{d \in D} q_l^{c},\; 1 \leq c \leq C_l,
\end{equation} 
where $D$ is the training dataset with size $|D|$.

Finally, according to the compression ratio $\mathcal{V}_k$, the original output tensor of CL $l$ can be compressed by pruning the less important channels and, thus, it outputs the compressed intermediate tensor $\hat{X}_l \in \mathbb{R}^{C'_l \times H_l \times W_l}$. Note that the number of channels of the compressed tensor is $C'_l = C_l/\mathcal{V}_k$.
\subsection{Feature recovery module}
In a well-trained CNN, there exist many redundant and similar features. Since the computational operation of CNN is essentially a series of linear and nonlinear transformations, some redundant features can be obtained from other features by performing inexpensive nonlinear transformation operations. Motivated by this, we design a light-weight CNN-based FR module to recover the intermediate tensor of CL $l$. 

\begin{figure}
    \centering
    \includegraphics[width=0.35\textwidth]{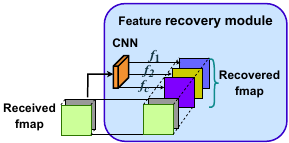}
    \caption{The feature recovery module of the Decoder.}
    \label{fig:decoder}
    \vspace{-3mm}
\end{figure}

As entropy coding is lossless, the entropy decoding yields the original compressed intermediate tensor $\hat{X}_l$. Therefore, we just need to generate the channels pruned by the CA module using $\hat{X_l}$, thereby rebuilding the intermediate tensor $X_l$, as shown in Fig. \ref{fig:decoder}. Unlike previous work, we use all the channels of the received tensor to generate each pruned channel, which allows better learning and recovery of the representation of the channels that are pruned. 

To illustrate the feature recovery module, we use a function $f_c\left(\cdot\right)$ to represent the computation operation of learning the $c \, $th channel pruned by the CA module. Thus, the recovered $c \, $th channel can be denoted as
\begin{equation}
    X_l^c = f_c \left(\hat{X}_l\right), c \leq C_l-C'_l
\end{equation} 
where the recovered $C_l-C'_l$ channels will be concatenated to the received tensor $\hat{X}_l$ as the input for CL $l+1$.

\subsection{Training strategy}
The proposed AECNN architecture can be trained in an end-to-end manner. However, this may result in very slow convergence. Therefore, we use a step-by-step training approach to train our proposed AECNN. We first insert the designed CA module into the original CNN model and then train the resulting neural network to figure out the importance of the channels. Based on the statistic of channels' importance, for a given compression ratio $\mathcal{V}_k$, $C_l(1-1/\mathcal{V}_k)$ channels with the lowest importance are identified as prunable. 
Then, we remove the inserted CA module and prune the original CNN model by removing the identified channels and the corresponding filters. Next, we fine-tune the pruned CNN model to recover the accuracy loss caused by the model pruning. Finally, we insert the designed FR module into the pruned CNN model and fine-tune the resulting CNN model to improve the inference accuracy. Throughout the training process, we do not consider the entropy encoding and decoding modules, because this lossless compression does not cause any accuracy loss. The detailed training process is described in Algorithm 1.
\begin{algorithm}
\caption{Training strategy for our proposed AECNN}
\KwIn{CNN model $\mathcal{G}$, the set of CLs $\mathcal{L}$, the set of compression ratio $\mathcal{V}$, the set of training data $\mathcal{D}$}
\KwOut{The set of AE-enhanced CNN models $\mathbf{G} = \left\{\mathcal{G}_1^{\mathcal{V}_1}, \cdots, \mathcal{G}_l^{\mathcal{V}_k}, \cdots,\mathcal{G}_{L-1}^{\mathcal{V}_K}\right\}$}
\For{$l =1,2,\cdots,L-1$}{
Insert CA module after the splitting point $l$.\\
Train the resulting network on training data $\mathcal{D}$.\\
Calculate the importance of each channel $c$ using (11).\\
Sort the importance of all the channels.\\
Remove the inserted CA module.\\
\For{$k = 1,2,\cdots,K$}{
Compress CL $l$ by pruning $C_l \left(1- 1/\mathcal{V}_k\right)$ less \\important channels.\\
Fine-tune the pruned CNN model.\\
Insert FR module into the pruned CNN model \\before CL $l+1$, then fine-tune the resulting\\  neural network to get $\mathcal{G}_l^{\mathcal{V}_k}$. 
}
}
\label{alg1}
\end{algorithm}
\begin{table*}
\centering
  \caption{Inference accuracy under different compression ratios of intermediate tensor and entropy encoding (measured on Raspberry pi 4B) and decoding (measured on RTX 2080TI) time}
\label{tab:BitErrors}
\scalebox{0.92}{
  \begin{tabular}{cccccccccccccccc}
    \toprule
 Splitting point &$C_l \times H_l \times W_l$ & $\mathcal{V}_k$ & BottleNet++ (\%)& CA\_Pruned (\%) &AECNN (\%) &$C'_l \times H_l \times W_l$  &Entropy (bit) & $t_{l,\mathcal{V}_k}^{\textit{enc}}$ (ms)  &$t_{l,\mathcal{V}_k}^{\textit{dec}}$ \\ 
    \midrule
 & & $2\times$ & $92.19(\pm0.22)$ &$95.58(\pm0.24)$  &$95.62(\pm0.20)$ &$32\times 56 \times 56$  &$11.13(\pm0.18)$ &4.53 &3.03\\ 
 & & $4\times$ & $\underline{\mathbf{91.66(\pm0.27)}}$ &$95.05(\pm0.26)$  &$\underline{\mathbf{95.30(\pm0.21)}}$ &$16\times 56 \times 56$  &$10.48(\pm0.16)$ &3.25 &2.41\\ 
 $l=1$ &$64\times 56 \times 56$ & $8\times$ & $91.49(\pm0.36)$ &$94.57(\pm0.11)$  &$94.68(\pm0.17)$ &$8\times 56 \times 56$  &$9.86(\pm0.21)$ &1.94 &1.52\\ 
 && $16\times$ & $90.92(\pm0.23)$ &$93.89(\pm0.54)$  &$93.98(\pm0.17)$ &$4\times 56 \times 56$  &$9.13(\pm0.27)$ &1.76 &1.37\\ 
 & & $32\times$ & $89.76(\pm0.18)$ &$92.69(\pm0.65)$  &$92.70(\pm0.31)$ &$2\times 56 \times 56$  &$8.62(\pm0.12)$ &1.03 &0.74\\ 
& & $\underline{\mathbf{64\times}}$ & $88.54(\pm0.44)$ &$91.10(\pm0.53)$  &$\underline{\mathbf{91.47(\pm0.25)}}$ &$1\times 56 \times 56$  &$\underline{\mathbf{7.73(\pm0.32)}}$ &0.62 &0.39\\ 
 & & $2\times$ & $93.28(\pm0.30)$ &$95.43(\pm0.19)$  &$95.48(\pm0.33)$ &$128\times 56 \times 56$  &$12.16(\pm0.27)$ &7.89 &4.76 \\ 
 & & $4\times$ & $92.25(\pm0.17)$ &$95.23(\pm0.25)$  &$95.24(\pm0.28)$ &$64\times 56 \times 56$  &$11.51(\pm0.29)$ &5.07 &3.21\\ 
 $l=2$ &$256\times 56 \times 56$ & $8\times$ & $91.70(\pm0.39)$ &$94.96(\pm0.14)$  &$95.05(\pm0.25)$ &$32\times 56 \times 56$  &$10.99(\pm0.24)$ &3.51 &2.60\\ 
 && $16\times$ & $91.64(\pm0.31)$ &$94.80(\pm0.30)$  &$94.85(\pm0.16)$ &$16\times 56 \times 56$  &$10.42(\pm0.21)$ &3.00 &2.35 \\ 
 & & $32\times$ & $90.86(\pm0.28)$ &$94.61(\pm0.21)$  &$94.64(\pm0.19)$ &$8\times 56 \times 56$  &$9.82(\pm0.18)$ &1.87 &1.46\\ 
& & $64\times$ & $90.63(\pm0.22)$ &$93.63(\pm0.28)$  &$93.78(\pm0.19)$ &$32\times 56 \times 56$  &$9.19(\pm0.22)$ &1.90 &1.44\\ 
  \bottomrule
\end{tabular}}
\vspace{-4mm}
\end{table*}

\section{Performance Evaluation}\label{simulation}
\subsection{Experimental Setup}
We consider the classification task of the Caltech-101 dataset \cite{caltech101}, which consists of about 9,000 images divided into 101 categories. Each category contains about 40 to 800 images with varying resolutions from $200\times 200$ to $300 \times 300$ pixels. Since IoT devices typically collect and process images at different resolutions, the Caltech-101 dataset is well suited for simulating conditions in IoT applications. In this experiment, we consider a device-edge co-inference system consisting of an IoT device (Raspberry pi 4B) and an ES (RTX 2080TI) to perform image classification based on the popular ResNet-50 \cite{He_2016_CVPR}. To perform the co-inference, we split ResNet-50 at different splitting points and compress the intermediate tensor with different compression ratios $\mathcal{V} = \left\{2,4,8,16,32,64\right\}$. Since ResNet-50 introduces a branching structure with residual blocks instead of a sequential structure, the first CL and each residual block are considered as the candidate splitting points in our evaluation. To validate the effectiveness of our proposed AECNN, we compare it with the state-of-the-art work BottleNet++ \cite{BottleNet++}, which encodes the intermediate tensor by directly resizing its dimension.

\addtolength{\topmargin}{0.03in}

\subsection{Importance of channels}
\begin{figure}
    \centering
    \includegraphics[width=0.4\textwidth]{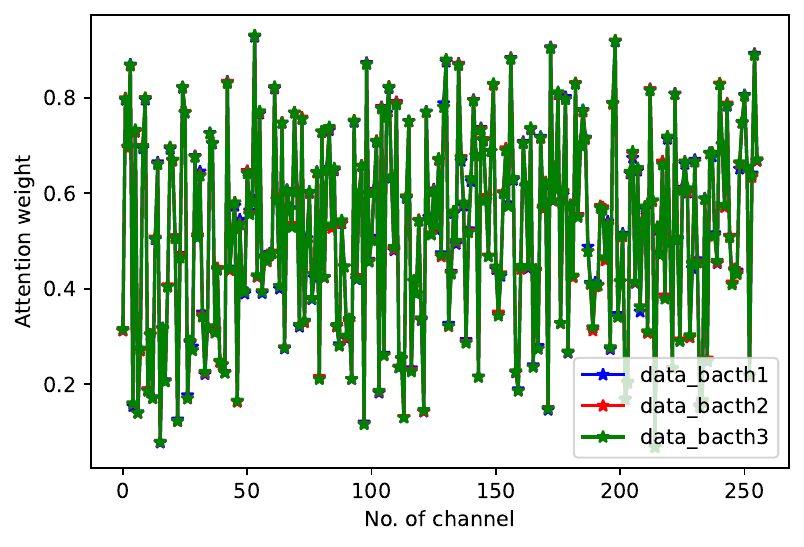}
    \vspace{-2mm}
    \caption{Attention weights generated from different batches of input data when splitting the model at $l=1$.}
    \label{fig:attention}
    \vspace{-4mm}
\end{figure}

To verify the robustness of the statistical method for calculating the importance of channels, we use the same amount of input data from different batches to calculate the importance of each channel. 
Due to the page limitation, we thereby take the first candidate point as an example and plot the channels' importance of the intermediate tensor by randomly sampling three batches of input data, as shown in Fig. \ref{fig:attention}. 
Even though the importance of the channels vary depending on the input data, the overall trend of the importance of channels calculated from these three batches of data is essentially consistent, which demonstrates the feasibility of the statistical method we used for calculating the importance of channels.

\subsection{Performance evaluation and analysis}
In a device-edge co-inference system, the latency is mainly caused by the computation and communication time on the IoT device. Therefore, we should split the CNN model as early as possible (near the input layer) to reduce the computation on the IoT device and compress the intermediate tensor as much as possible without compromising too much accuracy. In our experiment, we found that the computation time on the IoT device alone exceeds 100ms, if ResNet-50 is split at the third or later candidate points, which is not suitable for real-time inference. Therefore, we mainly consider the first and second candidate points. In our experiments, the inference accuracy of the original ResNet-50 on the test dataset is $95.84(\pm0.35) \%$. 
\subsubsection{Inference accuracy under different compression ratios}
Table I compares the inference accuracy of AECNN and BottleNet++ at different compression ratios of the intermediate tensor, in which ``CA\_Pruned” means the pruned ResNet-50 without FR module. We can see that AECNN improves the accuracy of CA\_Pruned ResNet-50, which demonstrates the effectiveness of the proposed FR module. In general, higher compression
results in more accuracy loss due to the lack of comprehensive presentation of the
features, but BottleNet++ suffers more accuracy loss
than AECNN at higher compression. Regardless of the spitting point and the compression ratio, AECNN can achieve higher accuracy than BottleNet++. For example, when $l=1$ and $\mathcal{V}_k=4$, AECNN can improve the accuracy from 91.66\% to 95.30\%. Moreover, with the same communication overhead, a higher accuracy is achieved by splitting the model at the first splitting point than the second. For example, splitting the model at the first point achieve the accuracy of 93.98\% when $\mathcal{V}_k=16$, while that of the second point is 93.78\% when $\mathcal{V}_k=64$. Note that in this case, the compressed data size at the first point is $4\times64\times56\times56/16$, which is equivalent to $4\times256\times56\times56/64$ at the second point.
Therefore, splitting the model at the first point is a better option, in addition, it uses less computation time and can reduce
the overall task completion time.
At the first splitting point, AECNN can compress the intermediate tensor by more than $256 \times$ (i.e., $64 \times 32/7.73$) using channel pruning and entropy coding, with accuracy loss of only about 4\% (i.e., $95.84\%-91.47\%$). 
\begin{figure*}
    \centering
    \subfigure[]{
        \label{fig:performancea}
    \includegraphics[width=0.35\textwidth]{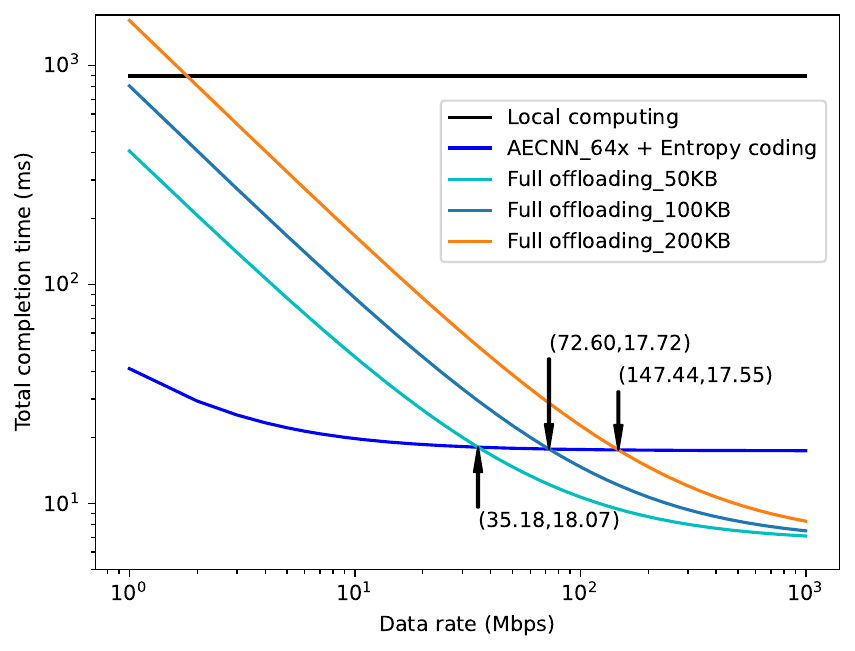}}
    \hspace{15mm}
    \subfigure[]{
    \includegraphics[width=0.35\textwidth]{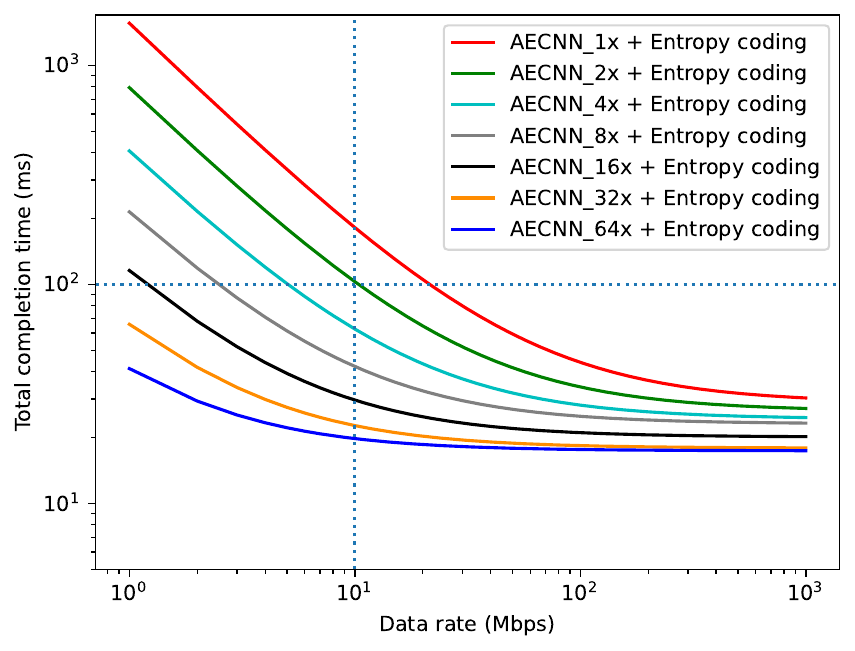}
        \label{fig:performanceb}}
        \vspace{-2mm}
    \caption{Task completion time under (a) different sizes of input images and (b) different compression ratios $\mathcal{V}_k$ }
    \vspace{-5mm}
\end{figure*}
\subsubsection{Task completion time under time-varying wireless channel states}
Since the offloading channel state is stochastic in practice, we calculate the task completion time of AECNN at various transmission data rates. In Fig. \ref{fig:performancea}, we plot the completion time of (i) AECNN splitting at the first point plus entropy coding, (ii) local computing and (iii) full offloading, for images with different input sizes. Note that entropy coding will not be used, when the reduced communication time is less than the introduced entropy encoding and decoding time. In addition, the communication time of AECNN and local computing are not affected by the size of the input image. It is clear that the completion time of local computation is constant, independent from the transmission data rate, while that of our proposed method and full offloading decreases with increasing transmission data rate. Compared to local computing, the proposed method and full offloading can perform inference faster. Compared to full offloading, AECNN can perform inference faster when the transmission data rate is low. For images of $1920\times1080$ pixels (i.e. 200 KB), AECNN outperforms the full offloading as long as transmission data rate is below 147.44 Mbps. In Fig. \ref{fig:performanceb}, we plot the completion time of AECNN at different compression ratios, where "$1\times$" indicates no compression. In general, high compression can accelerate inference but at the cost of accuracy loss. To trade off the latency and accuracy for time-constrained IoT applications, Fig. \ref{fig:performanceb} can be used to select the optimal compression ratio which gives the best possible accuracy within the time constraint. For example, when the transmission data rate is 10 Mbps, the compression ratio $\mathcal{V}_k = 4$ is the best among the pre-defined options for the IoT applications with the latency requirement of 100 ms.

\section{Conclusion}\label{conclusion}
In this paper, we studied the CNN inference task offloading in a device-edge co-inference system. Inspired by semantic communication, we proposed AECNN to extract the importance information of intermediate CNN features. We designed a CA module to figure out the channels' importance, then compressed the intermediate tensor by pruning the channels with lower importance. We used entropy encoding to encode the compressed intermediate tensor to further reduce communication time. In addition, we designed a lightweight CNN-based FR module to recover the intermediate tensor through learning from the received compressed tensor to improve accuracy. We used a step-by-step approach to fasten the training of the resulting neural networks of ResNet-50. The experimental results show that AECNN can compress the intermediate tensor by more than $256 \times$ thereby significantly reducing the communication time with only about 4\% accuracy loss, which outperforms the state-of-the-art work, BottleNet++. Compared to full offloading, AECNN can achieve shorter completion time, in particular, under poor channel condition. In our future research, we will further apply AECNN to multi-access edge computing networks.

\ifCLASSOPTIONcaptionsoff
  \newpage
\fi

\section*{Acknowledgment}
This work is supported by Agile-IoT project (Grant No. 9131-00119B) granted by the Danish Council for Independent Research.

\bibliographystyle{IEEEtran}
\bibliography{ddnn}

\end{document}